\documentclass[final]{cvpr}

\usepackage{times}
\usepackage{epsfig}
\usepackage{graphicx}
\usepackage{amsmath}
\usepackage{amssymb}
\usepackage{makecell}
\usepackage{array}
\usepackage{float}
\usepackage{placeins}
\usepackage[skip=2pt]{caption}
\usepackage{multirow}

\usepackage[pagebackref=true,breaklinks=true,colorlinks,bookmarks=false]{hyperref}

\def\cvprPaperID{****} % *** Enter the CVPR Paper ID here
\def\confYear{CVPR 2021}

\begin{document}

%%%%%%%%% TITLE
\title{Background Noise Reduction of Attention Map for Weakly Supervised Semantic Segmentation}

\author{Izumi Fujimori\textsuperscript{\rm 1}, Masaki Oono\textsuperscript{\rm 1}, Masami Shishibori\textsuperscript{\rm 1}\\
\textsuperscript{\rm 1}Tokushima University\\
% Institution1 address\\
{\tt\small \{Izumi Fujimori\}c612435027@tokushima-u.ac.jp} \\
%{\tt\small zhibozhang@cs.toronto.edu, j.jang@lgresearch.ai, ssanner@mie.utoronto.ca} \\
}

\maketitle
%%%%%%%%% ABSTRACT
\begin{abstract}
	In weakly-supervised semantic segmentation (WSSS) using only image-level class labels, 
	a problem with CNN-based Class Activation Maps (CAM) is that they tend to activate the most discriminative local regions of objects. 
	On the other hand, methods based on Transformers learn global features but suffer from the issue of background noise contamination.
	This paper focuses on addressing the issue of background noise in attention weights within the existing WSSS method based on Conformer, known as TransCAM.
	The proposed method successfully reduces background noise, leading to improved accuracy of pseudo labels. 
	Experimental results demonstrate that our model achieves segmentation performance of 70.5\% on the PASCAL VOC 2012 validation data, 
	71.1\% on the test data, and 45.9\% on MS COCO 2014 data, outperforming TransCAM in terms of segmentation performance.
\end{abstract}

%%%%%%%%% BODY TEXT
%%%Introduction%%%
\section{Introduction}
Semantic Segmentation is a fundamental task in computer vision, 
becoming increasingly important in a wide range of applications such as autonomous driving and land cover classification. 
While traditional supervised semantic segmentation methods have yielded excellent results\cite{badrinarayanan2017segnet}\cite{long2015fully}, 
they demand extensive labeled data for training, making annotation efforts a significant hurdle. 
To tackle this challenge, Weakly-Supervised Semantic Segmentation (WSSS) has emerged. 
WSSS aims to reduce annotation costs by utilizing annotation information other than pixel-level annotations. 
Various methods have been proposed in the realm of WSSS, including bounding boxes\cite{lee2021bbam}, scribbles\cite{lin2016scribblesup}, and points\cite{bearman2016s}. 
Among these, the most cost-effective approach involves utilizing only image-level class labels. 
This method leverages label information from predicted classes for the entire image to assign class labels to individual pixels. 
Image-level class labels are easily obtainable from existing large-scale datasets such as ImageNet\cite{russakovsky2015imagenet}, substantially reducing annotation costs. 
This paper focuses on WSSS using image-level class labels and proposes a method to enhance its segmentation accuracy.
\begin{figure}[htbp!]
    \centering
    \includegraphics[width=0.4\textwidth, height=6cm, bb=80 0 870 700]{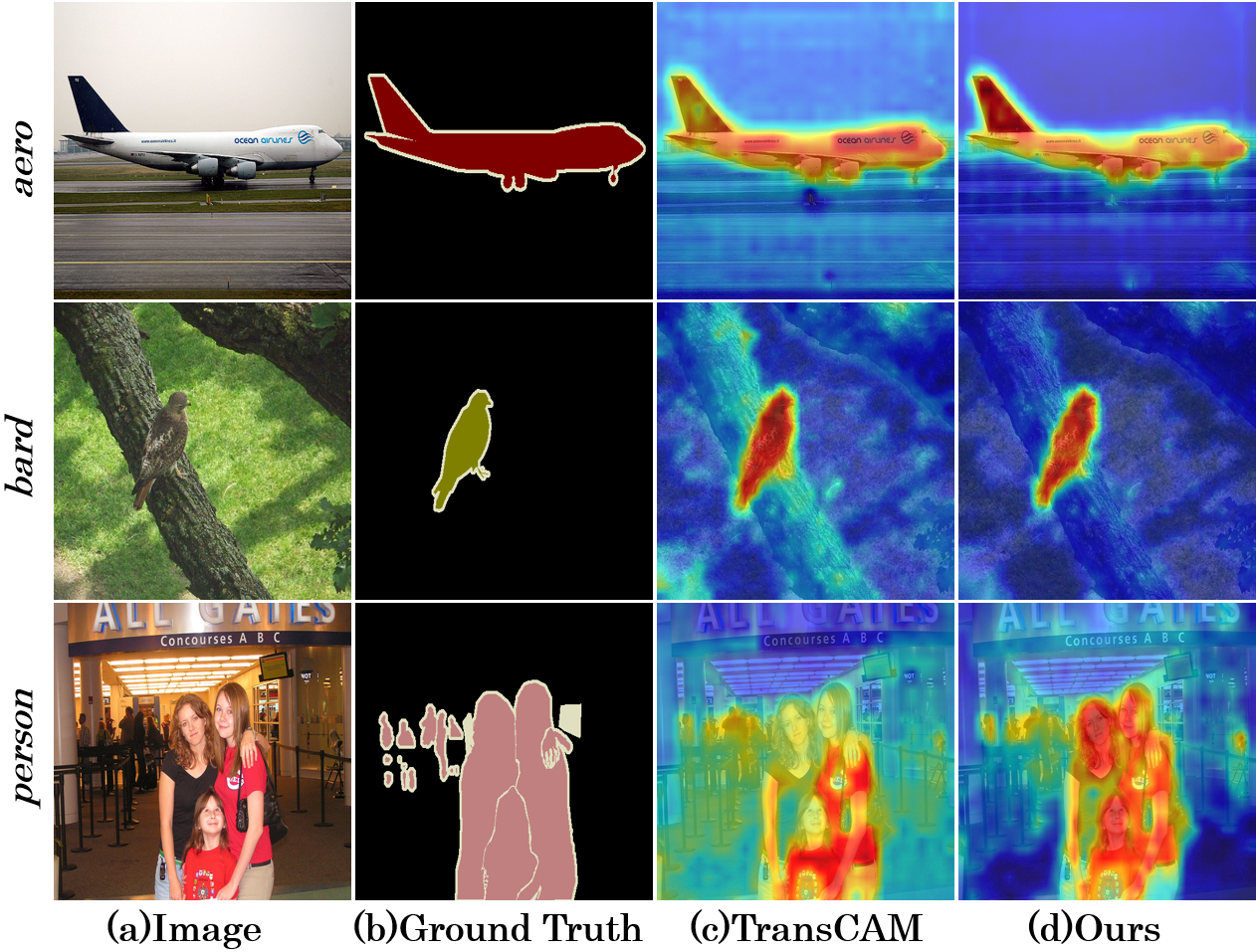}
    \caption{Comparison of CAMs generated by TransCAM and our proposed method. (a)Input image, (b)Ground Truth, (c)TransCAM, (d)CAMs by our method.}
    \label{fig:intro}
\end{figure}

In recent years, methods for WSSS have seen remarkable advancements. 
A common approach in WSSS involves generating pseudo labels using Class Activation Maps\cite{zhou2016learning} (CAM) based on Convolutional Neural Networks (CNNs). 
However, since CNNs learn features locally, the generated CAMs tend to activate only parts of objects intensely.
Vision Transformer\cite{dosovitskiy2020image} (ViT), based on Transformer\cite{vaswani2017attention} models with Multi-Head Self-Attention (MHSA), 
has brought breakthroughs in image recognition tasks. It has also been applied to WSSS, 
with methods utilizing attention maps to represent global features. 
However, issues such as background noise contamination and mislabeling of classes arise. 
Background noise degrades the performance of CAMs.
Therefore, this approach proposes a method to reduce background noise. 
It is composed of Conformers\cite{peng2021conformer}, aiming to alleviate background noise in WSSS.
In the WSSS method based on Conformer called TransCAM\cite{li2023transcam}, 
CAM obtained from CNN is enhanced using attention maps obtained from ViT. 
During model training, both the CAM based on CNN and the class tokens output from ViT are fed into the loss function. 
In addition to CAM and class tokens, 
the proposed method inputs CAM enhanced with attention maps into the loss function to reduce background noise. 
By incorporating the CAM enhanced with attention maps into the loss function, 
the training progresses in a direction where background noise is reduced. 
Figure \ref{fig:intro} illustrates CAM generated by the proposed method compared to CAM generated by existing methods, 
showing a reduction in background noise in the proposed method.
In experiments using the PASCAL VOC 2012 dataset\cite{everingham2010pascal}, 
the segmentation performance was 70.5\% for the validation data and 71.1\% for the test data. 
For experiments using the MS COCO 2014 dataset\cite{lin2014microsoft}, the performance was 45.9\%. 
The main contributions of this study are as follows:
\begin{itemize}
 \item Proposing a method to input CAM enhanced with attention maps into the loss function during training, which reduces background noise derived from attention.
 \item Conducting experiments on PASCAL VOC 2012 and MS COCO 2014 datasets, achieving segmentation accuracy surpassing existing methods.
\end{itemize}

%%%Related Work%%%
\section{Related Work}
\subsection{Weakly supervised semantic segmentation based on CNN}
Research on generating pseudo labels in WSSS has been a critical endeavor for improving the accuracy of segmentation tasks. 
A common approach involves utilizing Class Activation Maps (CAMs) from CNN classification networks to generate pseudo labels, which are then used as training data for supervised segmentation learning. 
However, CAMs suffer from the issue of activating only localized regions of the target.
In recent years, research focusing on addressing the problem of localized activation in CAMs has advanced. 
SEC\cite{kolesnikov2016seed} extends the heat range of CAMs by expanding seed regions. 
Adversarial erasing\cite{wei2017object} removes highly discriminative regions of objects, activating the remaining less discriminative regions. 
SEAM\cite{wang2020self} applies consistency regularization to CAMs predicted from images subjected to affine transformations. 
Additionally, it proposes a Pixel Correlation Module to enhance CAM consistency. 
Hide-and-Seek\cite{kumar2017hide} randomly hides patches in images during training, prompting the network to explore other relevant parts. 
CPN\cite{zhang2021complementary} divides images into complementary patches for a more comprehensive seed region in CAMs. 
PSA\cite{ahn2018learning} learns a network to predict semantic affinity matrices between pairs of adjacent image coordinates and propagates semantic affinity matrices using random walk\cite{lovasz1993random}.
Despite many excellent improvements, CNN-based methods still have limitations in representing global features.
\subsection{Weakly supervised semantic segmentation based on Transformer}
The breakthrough achieved by Vision Transformer (ViT) in image recognition tasks has also been applied to WSSS. 
Excellent methods leveraging the advantage of Transformer in capturing global features have been proposed. 
TS-CAM\cite{gao2021ts} generates pseudo labels using patch tokens and attention maps. 
MCTformer+\cite{xu2023mctformer+} enhances class token discriminability by proposing the Contrastive Class Tokens Module, which multiplies class tokens. 
WeakTr\cite{zhu2023weaktr} estimates the importance of attention weights used to enhance CAMs and proposes the Adaptive Attention Fusion Module to allocate weights adaptively.
Efforts have also been made to utilize a Conformer\cite{peng2021conformer} with a dual-branch structure of CNN and ViT as the backbone network. 
TransCAM\cite{li2023transcam} is pioneering research that applies Conformer as the backbone network to the WSSS task, enhancing CAMs from the CNN branch with attention maps. 
MECPformer\cite{liu2023mecpformer} utilizes a Conformer backbone network and transforms images into complementary patches inspired by CPN\cite{zhang2021complementary}. 
It learns multiple estimation complementary patches across different learning epochs to reduce false detections.
He et al\cite{he2023mitigating} pointed out the phenomenon of excessive smoothing of attention weights causing irrelevant background noise and proposed the Adaptive Re-activation Mechanism With Affinity Matrix. 
By learning to make the affinity matrices derived from pseudo labels closer to the values of attention weights, excessive smoothing of attention weights is mitigated.
This paper proposes a method to alleviate background noise in attention maps by inputting CAMs enhanced with attention maps into the loss function for training.

%%%Methodology%%%
\section{Methodology}
\begin{figure*}[tb]
    \centering
    \includegraphics[width=0.34\textwidth, height=6cm, bb=370 -20 760 320]{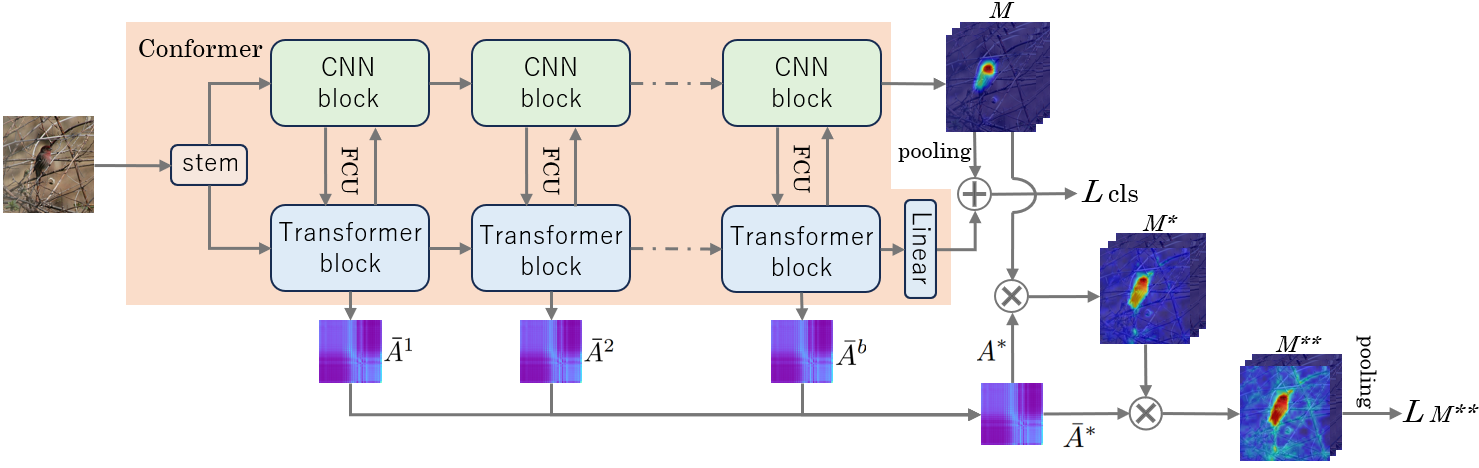}
    \caption{Overview of the proposed method. Create TransCAM from input images. Add noise to TransCAM and input it to the loss function. 
	Train TransCAM using image-level lable.  
	Attention map is explicitly incorporated into the training process to suppress the background noise derived from the Attention map. 
	Pseudo-labels are created from the initial seed obtained from TransCAM. The trained TransCAM has reduced background noise.}
    \label{fig:proposed method}
\end{figure*}

Figure \ref{fig:proposed method} illustrates an overview of the proposed method. 
The backbone network is composed of the Conformer architecture. 
Refined feature maps enhanced by attention maps are augmented with the noise $\bar{A}^{*}$. 
Next, the feature maps with added noise are fed into the loss function. 
Pseudo-labels are generated from seeds with reduced background noise.

%3.1
\subsection{Conformer as the backbone} 
In this method, a Conformer serves as the backbone network. 
The Conformer adopts a hybrid structure comprising CNN-block and Transformer-block. 
The CNN-block is based on ResNet\cite{he2016deep}, while the Transformer-block is inspired by ViT\cite{dosovitskiy2020image}. 
The Conformer utilizes Feature Coupling Unit\cite{peng2021conformer} (FCU) to integrate the local features learned by the CNN-block with the global features learned by the Transformer-block.

%3.2
\subsection{Seed generation} 
%3.2.1
\textbf{CAM generation.} 
Let's denote the output of the last CNN-block as $f \in \mathbb{R}^{fc\times fh\times fw}$ , where $fc$, $fh$ and $fw$ represent the number of channels, height, and width of the feature map, respectively. 
The weight corresponding to the feature map of the final layer is represented as $w \in \mathbb{R}^{fc}$. 
We denote the CAM as ${M}\in \mathbb{R}^{S\times fh\times fw}$, where $S$ represents the number of classes. 
The CAM corresponding to the $s$-th class $M_s \in \mathbb{R}^{fh\times fw}$ can be formulated as follows, 
where $s$ represents a specific class:
\begin{equation}
	M_s=w^\top f
\end{equation}
The CAM generated from the CNN-block of the Conformer is designed to have a shape of $M_s \in \mathbb{R}^{\sqrt{N} \times \sqrt{N}}$,
where $N$ represents the size of the patch token. 

%3.2.2
\textbf{Attention map generation.} Let the attention weights obtained from the Transformer-block be represented as $A^{b,h}\in\mathbb{R}^{b\times h\times (1+N)\times (1+N)}$,
where $b$, $h$ and $1$ denote a specific block, a specific head, and the class token, respectively.
$A^{b,h}$ is computed using the following equation:
\begin{equation}
	A^{b,h}=\text{softmax}(\frac{Q^{b,h}K^{b,h\top}}{\sqrt{D/h}}) 
\end{equation}
In the above equation, $Q^{b,h}$ and $K^{b,h}$ represent the queries and keys obtained in the $b$-th Transformer block, and $D$ denotes the embedding dimensions of tokens. 
By averaging the attention weights $A^{b,h}$ across heads for each block, 
we obtain $\bar{A}^{b} \in \mathbb{R}^{b\times (1+N)\times (1+N)}$, computed as follows:
\begin{equation}
	\bar{A}^{b} = \frac{1}{H}\sum_{h}(A^{b,h})
	\label{fig:A}
\end{equation}
where $H$ represents the number of heads. 
Summing $\bar{A}^{b}$ across layers yields a single attention weight calculated from the attention weights of each block and each head.
This yields $A \in \mathbb{R}^{(1+N)\times (1+N)}$, computed as:
\begin{equation}
	A = \sum_{b}(\bar{A}^{b})
\end{equation}
In $A \in \mathbb{R}^{(1+N)\times (1+N)}$, removing the attention weights associated with the class token results in an attention map $A^{*} \in \mathbb{R}^{N\times N}$.

%3.2.3
\textbf{Refinement of CAM with Attention Map.} Reshaping the CAM corresponding to the $s$-th class $M_s \in \mathbb{R}^{\sqrt{N} \times \sqrt{N}}$ into ${M_s} \in \mathbb{R}^{N\times 1}$, 
enhanced by the attention map $A^{*} \in \mathbb{R}^{N\times N}$, is calculated as follows:
\begin{equation}
	{M_s}^{*} = A^{*} \cdot {M_s}
\end{equation}
where ${M_s}^{*}\in \mathbb{R}^{N\times 1}$, which is then reshaped into ${M_s}^{*}\in \mathbb{R}^{\sqrt{N} \times \sqrt{N}}$.
${M_s}^{*}$ is considered as the inference result. 
Consequently, ${M}^{*}\in \mathbb{R}^{S\times \sqrt{N}\times \sqrt{N}}$.

%3.3
\subsection{Background noise reduction and training}
Enhanced CAMs ${M}^{*}_s \in \mathbb{R}^{N\times 1}$, bolstered by attention maps, are not directly utilized in the loss function and thus do not contribute to the network's learning process. 
Consequently, it is presumed that ${M_s}^{*}$ may contain superfluous background noise. 
To address this, our method proposes monitoring ${M_s}^{*}$ by inputting it into the Multi-label Soft Margin Loss during training, thereby alleviating background noise. 
Furthermore, augmenting ${M_s}^{*}$ with noise $\bar{A}^{*}\in \mathbb{R}^{N\times N}$ during training proves to be effective.
The noise added to ${M_s}^{*}$ is denoted by ${M_s}^{**}$.
This noise addition is done as follows:
\begin{equation}
	{M_s}^{**} = \bar{A}^{*} \cdot {M_s}^{*}
\end{equation}
where, ${M_s}^{**}\in \mathbb{R}^{N\times 1}$, which is then reshaped into ${M_s}^{**}\in \mathbb{R}^{\sqrt{N} \times \sqrt{N}}$.
The noise $\bar{A}^{*} \in \mathbb{R}^{N\times N}$ is obtained by averaging the attention weights relevant to the class token in $\bar{A}^{b}$, 
calculated by averaging $\bar{A}\in \mathbb{R}^{(1+N)\times (1+N)}$ along the layer direction in Equation \ref{fig:A}. 
$\bar{A}$ is computed as follows: 
\begin{equation}
	\bar{A} = \frac{1}{B}\sum_{b}(\bar{A}^{b})
\end{equation}
where, $B$ represents the number of blocks.
$\bar{A}^{*}$ is only added as noise during training. 
By applying Global Average Pooling (GAP) to the CAM augmented with noise, ${M}^{**}\in \mathbb{R}^{S\times \sqrt{N}\times \sqrt{N}}$, 
we obtain $z\in \mathbb{R}^{S}$. We can then compute the multi-label soft margin loss $L_{M^{**}}$ for $z$ using the following equation:
\begin{eqnarray}
	L_{M^{**}}=-\frac{1}{S-1}\sum_{s=1}^{S-1}[{y_s}\ln(\frac{1}{1+\exp(-z_s)})\nonumber\\
	\qquad\qquad+(1-y_s)\ln(\frac{\exp(-z_s)}{1+\exp(-z_s)})]
	\label{fig:losscls}
\end{eqnarray}
where, $y_s\in\{0, 1\}$ represents the ground truth for class $s$, and $z_s\in \mathbb{R}$ is the classification prediction for class $s$. 
Additionally, we add the value obtained by applying a linear layer to ${M}\in \mathbb{R}^{S\times \sqrt{N}\times \sqrt{N}}$ after applying GAP and the class token, 
$t\in\mathbb{R}^{(1+N)\times D}$, 
and input the resulting value into Equation 8 to obtain the loss $L_{cls}$.
where, $D$ represents the embedding dimensions of tokens. 
We train using the following loss equation:
\begin{equation}
	L=L_{cls} + L_{M^{**}}
\end{equation}
During inference, $\bar{A}^{*}$ is not included as noise.
Inference follows the same method as TransCAM (Baseline).

%3.4
\subsection{Pseudo label generation}
Pseudo-label Generation Enhanced CAMs ${M}^{*}_s\in \mathbb{R}^{\sqrt{N}\times \sqrt{N}}$, strengthened by attention maps, 
are normalized within the range of $[0, 1]$. In this case, the class $s$ range excludes the background class, defined as $1\leq s \leq S-1$. 
We establish a hard threshold $ht = [0, 1]$. 
For each pixel in the normalized CAM ${M}^{*}\in \mathbb{R}^{S\times \sqrt{N}\times \sqrt{N}}$, 
we assign the class $s$ with the maximum value in the $s$ direction as the pseudo-label. 
Pixels with values below the Hard Threshold $ht$ are labeled as the background class, 
while those with values equal to or above $ht$ are classified as the foreground class.
%-------------------------------------------------------------------------

%%%Experiments%%%
\section{Experiments}
\subsection{Dataset and Evaluation Metric}
For the evaluation dataset, we utilized the PASCAL VOC 2012 and MS COCO 2014 datasets. 
In the PASCAL VOC 2012 dataset, segmentation was conducted into 21 classes, including a background class and 20 foreground classes. 
Although the training, validation, and test datasets contain 1,464, 1,449, and 1,456 images, respectively, it's common to augment the training data with 10,582 images from the Semantic Boundary Dataset\cite{hariharan2011semantic}. 
The MS COCO 2014 dataset involves segmentation into 81 classes, including a background class and 80 foreground classes, with 82,081 training images and 40,137 validation images. 
Evaluation on the training data was performed using 81,394 images with pixel-level ground truth labels. 
The mean Intersection over Union (mIoU) was employed as the evaluation metric.

\subsection{Implementation Details}
The experiments were conducted using the NVIDIA RTX A6000 (VRAM 48GB).
We utilized a Conformer pretrained on ImageNet-1k\cite{russakovsky2015imagenet} as the backbone network. AdamW\cite{loshchilov2017decoupled} was employed as the optimizer with a learning rate of $5e^{-5}$ and a weight decay of $5e^{-4}$. 
During training, images were randomly resized within the range of $[320, 640]$ and then cropped to a size of $512\times 512$. 
During inference, images resized to $256\times256$, $512\times512$, and $768\times768$ were input to the model. 
CAMs were generated at each scale, which were then resized to the original size and aggregated. 
By utilizing multi-scale processing, predictions with high mIoU accuracy could be achieved. 
For training on the PASCAL VOC 2012 dataset, we set the number of epochs to 22 and the batch size to 8. 
For the MS COCO 2014 dataset, we set the number of epochs to 10 and the batch size to 16. 
PSA\cite{ahn2018learning} was employed for post-processing of pseudo-labels. 
The pseudo-labels generated after post-processing were refined using CRF\cite{krahenbuhl2011efficient}. 
The supervised segmentation model was based on ResNet38\cite{wu2019wider} and DeepLab\cite{chen2014semantic}.

\subsection{Experimental Results}
\begin{table}[htbp]
	\caption{Comparison of mIoUs (\%) for seed maps and pseudo-semantic masks of Baseline and the proposed method on the train set.}
	\centering
	\begin{tabular}[t]{l c c}
	  \hline
	  Method&VOC&COCO\\
	  \hline
	  \hline
	  TransCAM\cite{li2023transcam}\textup{(Baseline)}&64.9&46.2\\
	  Ours&66.2&47.4\\
      \hline
      TransCAM\cite{li2023transcam}\textup{(Baseline)}+ PSA\cite{ahn2018learning}&70.2&48.7\\
	  Ours + PSA\cite{ahn2018learning}&71.3&50.6\\
      \hline
	\end{tabular}
    
	\label{tab:seed}
\end{table}
\begin{figure*}[htbp]
    \centering
    \includegraphics[width=0.34\textwidth, height=6cm, bb=310 -20 640 320]{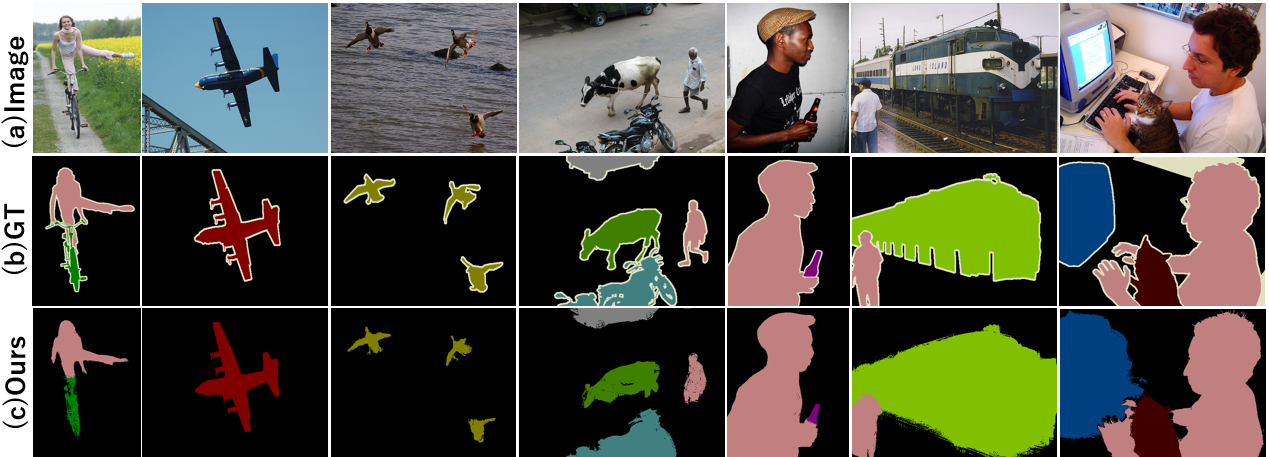}
    \caption{Qualitative segmentation results on the PASCAL VOC 2012 validation set. 
	(a) Original images; (b) Ground truth; (c)Prediction of Ours.}
    \label{fig:pseudo label}
\end{figure*}

\begin{table*}[tb]
	\caption{Performance of Categorical Semantic Partitioning on val set and test set in PASCAL VOC 2012.}
	\label{table:data_type}
	\centering
	\resizebox{\textwidth}{!}{
		\begin{tabular}[t]{l c c c c c c c c c c c c c c c c c c c c c c}
			\hline
			Method&bkg&aero&bike&bird&boat&bottle&bus&car&cat&chair&cow&table&dog&horse&mbk&person&plant&sheep&sofa&train&tv&mIoU\\
			\hline
			TransCAM(Baseline)&91.3&81.9&35.4&84.7&67.6&67.9&\textbf{87.5}&80.5&86.5&31.4&73.9&52.5&84.0&74.9&74.6&79.0&44.7&84.1&47.0&\textbf{78.4}&\textbf{46.6}&69.3\\
			ours&\textbf{91.6}&\textbf{85.3}&\textbf{35.8}&\textbf{87.1}&\textbf{70.5}&\textbf{72.2}&87.0&\textbf{82.8}&\textbf{87.7}&\textbf{32.5}&\textbf{77.2}&\textbf{52.6}&\textbf{84.0}&\textbf{75.9}&\textbf{74.7}&\textbf{81.1}&\textbf{50.9}&\textbf{85.1}&\textbf{48.4}&76.5&42.1&\textbf{70.5}\\
			\hline
			TransCAM(Baseline)&91.2&82.2&36.2&89.8&57.8&65.1&85.7&81.5&84.2&28.3&77.4&55.9&82.1&80.0&78.9&76.5&48.0&84.7&57.2&72.7&45.8&69.6\\
			ours&\textbf{91.6}&\textbf{85.0}&36.1&89.7&\textbf{58.9}&\textbf{67.9}&\textbf{86.1}&\textbf{82.8}&\textbf{86.8}&\textbf{30.6}&\textbf{81.0}&\textbf{60.5}&\textbf{82.6}&\textbf{81.3}&\textbf{79.7}&\textbf{79.4}&\textbf{56.9}&\textbf{85.5}&\textbf{59.3}&71.2&41.1&\textbf{71.1}\\
			\hline
		\end{tabular}
	}
	
	\label{tab:voc_pseudo}
\end{table*}

We evaluated the accuracy of initial pseudo-label assignment. 
Table \ref{tab:seed} presents the experimental results conducted on the PASCAL VOC 2012 training data and MS COCO 2014 training data. 
For the PASCAL VOC 2012 training data, the proposed method achieved 66.2\%, surpassing the Baseline accuracy by 1.3 percentage points. 
On the MS COCO 2014 training data, the proposed method achieved 47.4\%, surpassing the Baseline accuracy by 1.2 percentage points.

\begin{table}[tb]
	\caption{Comparison of mIoUs (\%) of segmentation performance in PASCAL VOC. 
	Backbone indicates network for semantic segmentation.
	I denotes image-level labels and S denotes saliency maps.}
	\label{table:data_type}
	\small
	\centering
	\begin{tabular}[t]{l c c c c }
	  \hline
	  Method&Backbone&Sup.&\textit{val}&\textit{test}\\
	  \hline
	  AuxSegNet\cite{xu2021leveraging}\tiny{$\mathrm{ICCV21}$}&RN38&I+S&69.0&68.6\\
      L2G\cite{jiang2022l2g}\tiny{$\mathrm{CVPR22}$}&RN38&I+S&72.0&73.0\\
	  MECPformer\cite{liu2023mecpformer}\tiny{$\mathrm{Arxiv23}$}&RN101&I+S&72.0&72.0\\
	  \hline
	  SEAM\cite{wang2020self}\tiny{$\mathrm{CVPR20}$}&RN38&I&64.5&65.7\\
      CONTA\cite{zhang2020causal}\tiny{$\mathrm{NIPS20}$}&RN38&I&66.1&66.7\\
      Kweon \textit{et al.}\cite{kweon2021unlocking}\tiny{$\mathrm{ICCV21}$}&RN38&I&68.4&68.2\\
      CDA\cite{su2021context}\tiny{$\mathrm{ICCV21}$}&RN38&I&66.1&66.8\\
	  CPN\cite{zhang2021complementary}\tiny{$\mathrm{ICCV21}$}&RN38&I&67.8&68.5\\
	  AdvCAM\cite{lee2022anti}\tiny{$\mathrm{CVPR21}$}&RN101&I&68.1&68.0\\
      AMN\cite{lee2022threshold}\tiny{$\mathrm{CVPR22}$}&RN101&I&70.7&70.6\\
      W-OoD\cite{lee2022weakly}\tiny{$\mathrm{CVPR22}$}&RN38&I&70.7&70.1\\
      SIPE\cite{chen2022self}\tiny{$\mathrm{CVPR22}$}&RN38&I&68.2&69.5\\
      Yoon \textit{et al.}\cite{yoon2022adversarial}\tiny{$\mathrm{ECCV22}$}&RN38&I&70.9&71.7\\
      OCR\cite{cheng2023out}\tiny{$\mathrm{CVPR23}$}&RN38&I&72.7&70.7\\
      LPCAM\cite{chen2023extracting}\tiny{$\mathrm{CVPR23}$}&RN38&I&72.6&72.4\\
      MCTformer+\cite{xu2023mctformer+}\tiny{$\mathrm{Arxiv23}$} &RN38&I&74.0&73.6\\
	  He \textit{et al.}\cite{he2023mitigating}\tiny{$\mathrm{Arxiv23}$}&RN38&I&69.9&70.0\\
	  \hline
	  TransCAM\cite{li2023transcam}\tiny{$\mathrm{JVCI23}$}&RN38&I&69.3&69.6\\
	  Ours&RN38&I&70.5&71.1\\
	  \hline
	\end{tabular}
	\label{tab:voc_val_test}
\end{table}

\begin{table}[tb]
	\caption{Comparison of mIoUs(\%) for segmentation performance on MS COCO val set.}
	\label{table:data_type}
	\small
	\centering
	\begin{tabular}[t]{l c c c}
	  \hline
	  Method&Backbone&Sup.&\textit{val}\\
	  \hline
	  AuxSegNet\cite{xu2021leveraging}\tiny{$\mathrm{ICCV21}$}&RN38&I+S&33.9\\
	  L2G\cite{jiang2022l2g}\tiny{$\mathrm{CVPR22}$}&RN101&I+S&44.2\\
	  MECPformer\cite{liu2023mecpformer}\tiny{$\mathrm{Arxiv23}$}&RN101&I+S&42.4\\
	  \hline
	  SEAM\cite{wang2020self}\tiny{$\mathrm{CVPR20}$}&RN38&I&31.9\\
	  CONTA\cite{zhang2020causal}\tiny{$\mathrm{NIPS20}$}&RN38&I&32.8\\
	  Kweon \textit{et al.}\cite{kweon2021unlocking}\tiny{$\mathrm{ICCV21}$}&RN38&I&36.4\\
	  CDA\cite{su2021context}\tiny{$\mathrm{ICCV21}$}&RN38&I&33.2\\
	  AdvCAM\cite{lee2022anti}\tiny{$\mathrm{CVPR21}$}&RN101&I&44.4\\
	  AMN\cite{lee2022threshold}\tiny{$\mathrm{CVPR22}$}&RN101&I&44.7\\
	  SIPE\cite{chen2022self}\tiny{$\mathrm{CVPR22}$}&RN38&I&43.6\\
	  Yoon \textit{et al.}\cite{yoon2022adversarial}\tiny{$\mathrm{ECCV22}$}&RN38&I&44.8\\
	  OCR\cite{cheng2023out}\tiny{$\mathrm{CVPR23}$}&RN38&I&42.5\\
	  LPCAM\cite{chen2023extracting}\tiny{$\mathrm{CVPR23}$}&RN38&I&42.8\\
	  MCTformer+\cite{xu2023mctformer+}\tiny{$\mathrm{Arxiv23}$}&RN38&I&45.2\\
	  \hline
	  TransCAM\cite{li2023transcam}\tiny{$\mathrm{JVCI23}$}&RN38&I&45.7\\
	  Ours&RN38&I&45.9\\
	  \hline
	\end{tabular}
	\label{tab:coco_val}
\end{table}

Next, we evaluated the performance of the pseudo-labels after post-processing. 
Table \ref{tab:seed} shows the experimental results conducted on the PASCAL VOC 2012 training data and MS COCO 2014 training data. 
For the PASCAL VOC 2012 training data, the proposed method achieved 71.3\%, surpassing the Baseline accuracy by 1.1 percentage points. 
On the MS COCO 2014 training data, the proposed method achieved 50.6\%, surpassing the Baseline accuracy by 1.9 percentage points.

We conducted supervised learning using pseudo-labels as the training data. 
Table \ref{tab:voc_pseudo} presents the semantic segmentation results on the class-wise PASCAL VOC 2012 validation data. 
The proposed method outperformed existing methods in 17 out of 20 classes, demonstrating the effectiveness of suppressing background noise not only for specific classes but also for each class individually. 
Figure \ref{fig:pseudo label} illustrates the inference results of the segmentation model trained with pseudo-labels generated by the proposed method, showing the original image, ground truth, and the segmentation results obtained by the proposed method, respectively.

Table \ref{tab:voc_val_test},\ref{tab:coco_val} presents the segmentation accuracy, where "Sup." in the middle column denotes image-level labels, and "S" represents saliency maps. 
For the PASCAL VOC 2012 validation data, the proposed method achieved 70.5\%, surpassing the Baseline accuracy by 1.2 percentage points. 
On the test data, it achieved 71.1\%, surpassing the Baseline accuracy by 1.5 percentage points. 
For the MS COCO 2014 validation data, it achieved 45.9\%, surpassing the Baseline accuracy by 0.2 percentage points.
\subsection{Investigating the Effects of Adding Noise}
In this section, we investigate the impact of varying levels of noise ($\bar{A}^{*}\in \mathbb{R}^{N\times N}$) on the inference results. 
We conducted training by sequentially adding noise ($\bar{A}^{*}$). 
Table \ref{tab:abl} compares the initial pseudo-label mIoU based on the amount of noise, obtained from the PASCAL VOC 2012 training data. 
From left to right, it shows the Baseline, results trained without noise, results trained with noise multiplied by 1 ($\bar{A}^{*}\cdot1$), and results trained with noise multiplied by 2 ($\bar{A}^{*}\cdot2$). 
The accuracy of the training results with noise was higher than that without noise, indicating the effectiveness of training with added noise. 
The decrease in accuracy of the training results without noise can be attributed to the suppression of both background noise and the activation of object regions in the attention map.
\begin{table}[htbp]
	\caption{Comparison of performance for different amounts of noise in the initial pseudo-labels in the PASCAL VOC train set.}
	\label{table:data_type}
	\centering
	\begin{tabular}[t]{l | c c c c}
	  \hline
	  noise&Baseline\cite{li2023transcam}&without noise&$\bar{A}^{*}\cdot1$&$\bar{A}^{*}\cdot2$\\
	  \hline
	  train&64.9&62.4&66.2&65.3\\
      \hline
	\end{tabular}
	\label{tab:abl}
\end{table}

\begin{figure}[htbp]
    \centering
    \includegraphics[width=0.6\textwidth, height=6cm, bb=0 0 560 300]{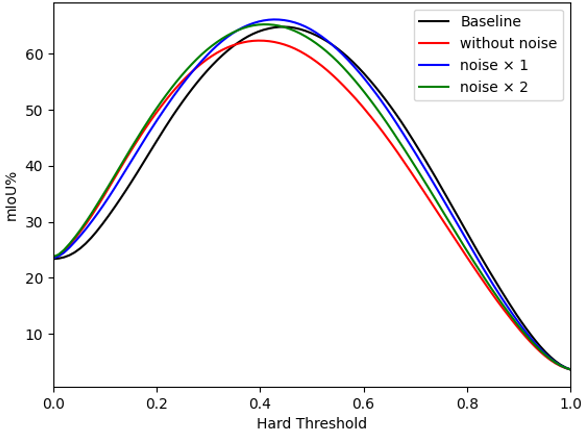}
    \caption{mIoU variation according to Hard thresholds}
    \label{fig:ht_miou}
\end{figure}
\begin{figure}[htbp]
    \centering
    \includegraphics[width=0.6\textwidth, height=6cm, bb=-20 0 630 370]{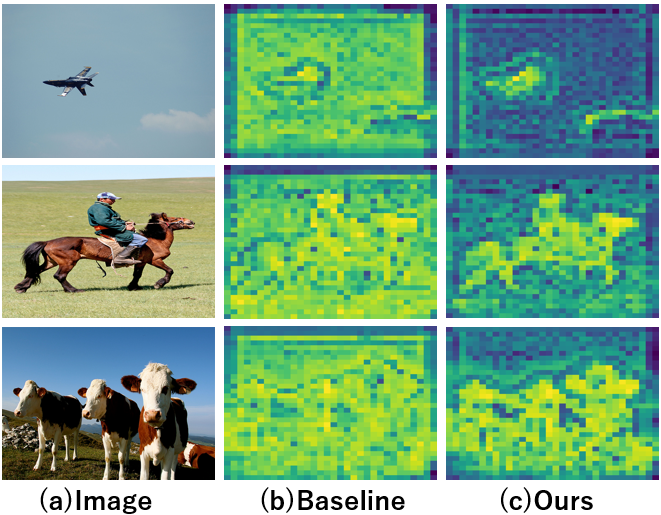}
    \caption{Comparisons of Attention map. (a) Original image; (b)Attention map generated by baseline; (c) Attention map generated by our method.}
    \label{fig:attention map}
\end{figure}
Figure \ref{fig:ht_miou} illustrates the transition of mIoU based on the Hard Threshold of pseudo-labels in PASCAL VOC 2012. 
The vertical axis represents mIoU, while the horizontal axis represents the Hard Threshold. 
The black line represents the transition of mIoU for pseudo-labels generated by the Baseline, while the red, blue, and green lines represent the transition of mIoU for pseudo-labels generated by the proposed method. 
When CAM contains a lot of background noise, at low Hard Thresholds, the background noise becomes part of the object area, leading to performance degradation. 
The proposed method exhibits higher mIoU when the Hard Threshold is lower than the Baseline, indicating a reduction in background noise.

Figure \ref{fig:attention map} compares the attention maps between the conventional method and the proposed method. 
It is evident that the activation of background regions is reduced in the attention maps generated by the proposed method compared to the conventional method.

\section{Conclusion}

This paper proposes a method to suppress background noise in attention maps by inputting attention map-enhanced CAMs into the loss function and evaluates its performance. 
The proposed method surpassed the accuracy of existing methods on the PASCAL VOC 2012 dataset and MS COCO 2014 dataset. 
Since the activation of object regions tended to be suppressed in the proposed method, future work could focus on designing loss functions aimed at suppressing background noise.

{\small
\bibliographystyle{ieee_fullname}
\bibliography{egbib.bib}
}

\end{document}